\pdfoutput=1

\documentclass[11pt]{article}

\usepackage[final]{acl}
\usepackage{graphicx}
\usepackage{times}
\usepackage{latexsym}
\usepackage{cite}
\usepackage{makecell}
\usepackage[T1]{fontenc}
\usepackage{booktabs}
\usepackage{amsmath}
\usepackage[utf8]{inputenc}

\usepackage{microtype}
\usepackage{inconsolata}
\usepackage{hyperref}
\usepackage{multirow}
\usepackage{subfig}
\usepackage{enumitem}
\usepackage{balance}
\usepackage{threeparttable}
\usepackage{amsmath,amsfonts,mathtools}
\usepackage{mathrsfs}
\usepackage{float}
\usepackage{enumitem}
\usepackage{arydshln}
\usepackage{graphicx}

\usepackage[linesnumbered,vlined,ruled]{algorithm2e}
\usepackage{enumitem}
\usepackage{graphicx}
\usepackage{subcaption}

\title{EFSA: Towards Event-Level Financial Sentiment Analysis}

\author{Tianyu Chen$^{1,2,3}$\footnotemark[1], Yiming Zhang$^{4}$\footnotemark[1], Guoxin Yu$^{1,2,3}$, Dapeng Zhang$^{5}$, Li Zeng$^{6}$\footnotemark[2], Qing He$^{1,2,3,4}$, Xiang Ao$^{1,2,3}$\footnotemark[2] \\
   $^{1}$Key Laboratory of AI Safety, Chinese Academy of Sciences~(CAS), Beijing, China. \\
   $^{2}$Key Lab of Intelligent Information Processing, Institute of Computing Technology, CAS, Beijing, China. \\
   $^{3}$University of Chinese Academy of Sciences, Beijing, China.\\
   $^{4}$Henan Institute of Advanced Technology, Zhengzhou University, Zhengzhou, China.\\
   $^{5}$School of IoT Engineering, Jiangsu Vocational College of Information Technology, Wuxi, China.\\
   $^{6}$Information Technology Department \uppercase\expandafter{\romannumeral1}, Shenzhen Stock Exchange. \\
  \texttt{\{chentianyu22s,aoxiang\}@ict.ac.cn} \\
  }

\begin{document}
\maketitle
\renewcommand{\thefootnote}{\fnsymbol{footnote}}
\footnotetext[1]{These authors contributed equally to this work.}
\footnotetext[2]{Correspondence to Xiang Ao and Li Zeng.}
\begin{abstract}
In this paper, we extend financial sentiment analysis~(FSA) to event-level since events usually serve as the subject of the sentiment in financial text. Though extracting events from the financial text may be conducive to accurate sentiment predictions, it has specialized challenges due to the lengthy and discontinuity of events in a financial text. To this end, we reconceptualize the event extraction as a classification task by designing a categorization comprising coarse-grained and fine-grained event categories. Under this setting, we formulate the \textbf{E}vent-Level \textbf{F}inancial \textbf{S}entiment \textbf{A}nalysis~(\textbf{EFSA} for short) task that outputs quintuples consisting of (company, industry, coarse-grained event, fine-grained event, sentiment) from financial text. A large-scale Chinese dataset containing $12,160$ news articles and $13,725$ quintuples is publicized as a brand new testbed for our task. A four-hop Chain-of-Thought LLM-based approach is devised for this task. Systematically investigations are conducted on our dataset, and the empirical results demonstrate the benchmarking scores of existing methods and our proposed method can reach the current state-of-the-art. Our dataset and framework implementation are available at \url{https://github.com/cty1934/EFSA}.

\end{abstract}

\section{Introduction}

Since Robert F. Engle was awarded the Nobel Prize because he researched the influence of news on financial market volatility~\citep{engle1993measuring}, it catalyzed a surge in academic interest in Financial Sentiment Analysis~(FSA)~\citep{luo2018beyond,xing2020financial,du2024financial}. FSA holds significant importance within the application domain of sentiment analysis~\citep{du2024financial}, encompassing the study of financial textual sentiment~\citep{kearney2014textual} within news to forecast financial market dynamics. Recall that it is the events described in the financial texts and the related sentiments that dominate the impact of financial news on market volatility~\citep{xing2020financial}, the primary focus of FSA should sit on events extraction and related sentiment analysis. 

\begin{figure}[t]
    \centering
    \includegraphics[width=1\linewidth]{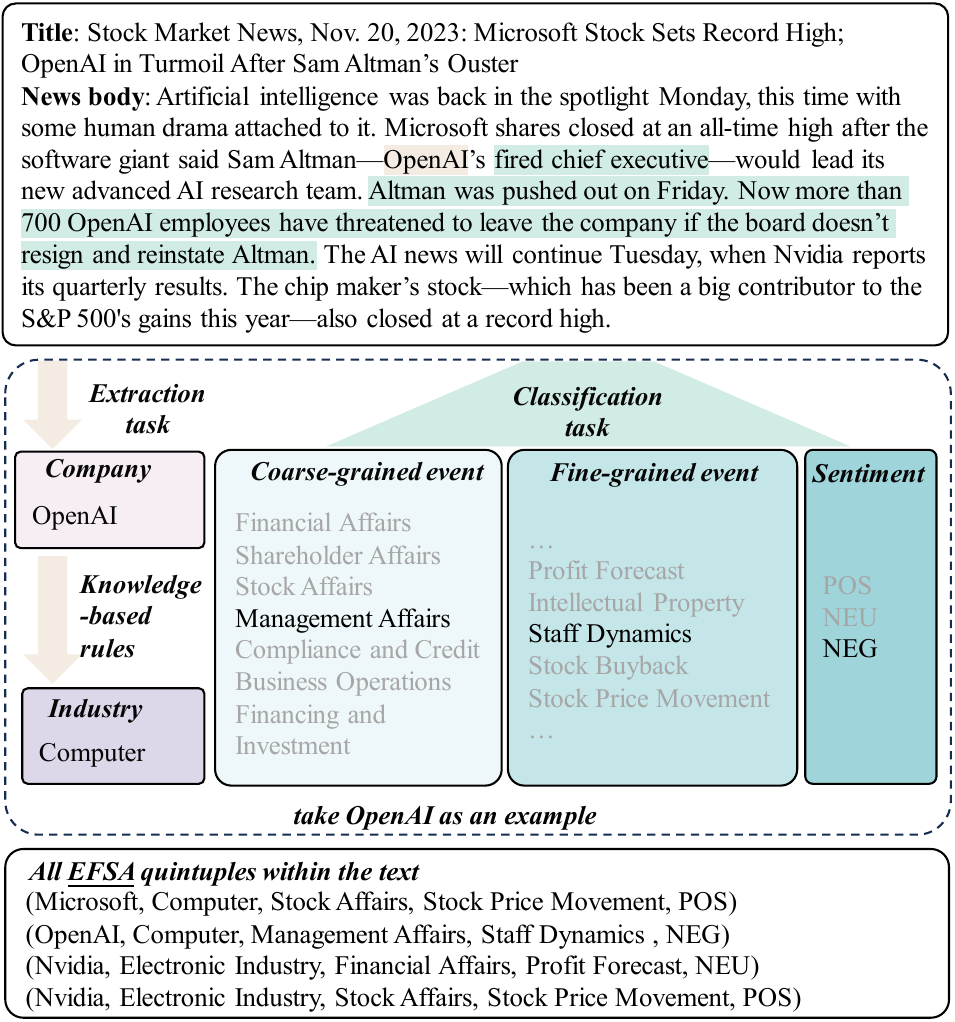}
    \caption{An example of Event-level Financial Sentiment Analysis from financial news. There could be multiple entities associated with their events with different sentiments.}
    \label{fig:1}
\end{figure}

However, most existing FSA studies focus on predicting entities and sentiments while neglecting the analysis of events within financial texts. To name some, FiQA~\citep{maia201818}, an open challenge financial news dataset, is designed to analyze sentiment corresponding to a certain entity. SEntFiN~\citep{sinha2022sentfin} is a news headline dataset for entity analysis, including entity recognition from a predefined list and related sentiment analysis. A most recent FSA benchmark,  namely Fin-Entiy~\citep{tang2023finentity}, aims to jointly predict the entities and the associated sentiments from financial news. 
Nevertheless, financial text's sentiments frequently link to particular events. 
For example, in Figure~\ref{fig:1}, the same entity \textit{Nvidia} exhibits distinctly opposite sentiments due to two different events: \textit{stock price movement} and \textit{profit forecast}. From this example, we can be aware that the event usually serves as the subject of the sentiment in financial text, while the entity is the target of the emotional impact. Extracting events from financial text may be conducive to accurate sentiment predictions.

To this end, we aim to discover the events in financial text to provide an easier financial sentiment prediction. 
An intuitive way to identify an event from financial text is extracting a specific text span from the original text just like existing aspect-based sentiment analysis~(ABSA) tasks that find both aspect and opinion terms from customer's comments~\citep{zhang2022survey,yu2023making,yu2021self}. 
However, direct adapting approaches of ABSA for this task could be ineffective even infeasible due to the events in the financial text being overlong and discontinuous~(c.f. Figure~\ref{fig:1}). We will provide a further discussion about the similarities and differences between ABSA and our task in Section~\ref{Section 2.3}.


In this paper, based on our observations, we provide an alternative setting for FSA and propose a novel task, named \textbf{E}vent-Level \textbf{F}inancial \textbf{S}entiment \textbf{A}nalysis (\textbf{EFSA}), involving the prediction of quintuples~(company, industry, coarse-grained event, fine-grained event, sentiment). Here we have enhanced the FSA tasks in two facets. First, to overcome the difficulties associated with extracting events from financial texts, we reconceptualize the event extraction task as a classification task. We design a categorization system that comprises both coarse-grained and fine-grained event categories, specifically tailored for various event types in financial news. Besides, we construct a knowledge-based rule to classify companies by industry, enabling FSA to elevate to higher dimensions, such as indices or sector markets. Our task setting can offer substantial practical value in financial applications, including stock trading, stock market anomaly attribution analysis, enterprise risk management, etc. Figure~\ref{fig:1} presents example quintuples in our EFSA task.

To support this task, we annotated a large-scale dataset from Chinese financial news. The dataset includes $12,160$ news articles, selected from an initial set of over $50,000$ articles collected from mainstream Chinese financial news websites. Detailed annotations were conducted based on the above task settings. To the best of our knowledge, this dataset is a large-scale fine-grained annotated dataset for FSA and the largest Chinese dataset in the event-level FSA domain.

We conducted comprehensive experiments to benchmark our dataset. Empirical results demonstrate the EFSA task presents a significant challenge, even for advanced large language models~(LLMs) such as GPT-4, primarily due to the complexity of simultaneously predicting two categories and the fact that the sentiments in financial texts are primarily implicit. Recent studies on implicit sentiments underscore the efficacy of the Chain-of-Thought~(CoT) approach in reasoning implicit sentiments in fine-grained sentiment analysis~\citep{fei2023reasoning}. Consequently, we devise a framework utilizing a 4-hop Chain of Thought~(CoT) prompt based on LLMs, achieving the highest level of performance recorded for our task.

The main contributions of our work can be summarized as follows:
\begin{itemize}
    \item We propose a novel Event-Level Financial Sentiment Analysis task for financial sentiment analysis, named EFSA, including the prediction of quintuples consisting of~(company, industry, coarse-grained event, fine-grained event, and sentiment).
    \item We publicize the largest-scale Chinese financial corpus to support the EFSA task.
    \item We conduct systematically benchmark experiments to investigate the efficacy of existing methods for the EFSA task and introduce a novel LLM-based framework reaching the current state-of-the-art.
\end{itemize}

\section{Problem Formulation and Discussion}

In this section, we formulate the EFSA task and differentiate it from the traditional ABSA task.

\subsection{Problem Formulation}

The problem of EFSA is formulated as follows: Given an input financial news context \textit{x} contains \textit{n} words, EFSA aims to predict a set of quintuples \textit{(c, i, e}\textsuperscript{\textit{1}}, \textit{e}\textsuperscript{\textit{2}}, \textit{s)}, corresponding to~(\textit{company, industry, coarse-grained event, fine-grained event, sentiment polarity}). Here, \textit{company c}  is a text span within the sentence. Each \textit{company c} is categorized into a specific \textit{industry i} by knowledge-based rules.  \textit{Coarse-grained event e}\textsuperscript{\textit{1}}and \textit{fine-grained event e}\textsuperscript{\textit{2}} belong to two distinct predefined event type sets \textit{E}\textsuperscript{\textit{1}} and  \textit{E}\textsuperscript{\textit{2}}, where \textit{E}\textsuperscript{\textit{2}} is a finer-grained division of \textit{E}\textsuperscript{\textit{1}}. \textit{s} $\in$ \{positive, negative, neutral\} denotes the sentiment polarity.

\begin{table}
    \centering

    \begin{tabular}{ll} 
    \toprule
         Task & Output \\ 
         \midrule
         EFSA & $(c, i, e^{1}, e^{2}, s)$ \\  
         Coarse-grained EFSA & $(c, i, e^{1}, s)$ \\  
         Fine-grained EFSA & $(c, i, e^{2}, s)$ \\  
         Entity-Level FSA & $(c, i, s)$ \\ 
         \bottomrule
    \end{tabular}
\caption{The sub-tasks of EFSA.}\label{tab:1}
\end{table}

\subsection{The Sub-tasks of EFSA} 
The EFSA task could be further divided into two event-level sub-tasks, namely \textbf{C}oarse-grained EFSA~(C-EFSA) and \textbf{F}ine-grained EFSA~(F-EFSA), along with an entity-level FSA task. These tasks are outlined in Table~\ref{tab:1}. 
Due to the granularity difference of events categorization within different tasks, the difficulty of the three tasks, C-EFSA, F-EFSA, and complete EFSA, exhibits an increasing trend. This enables FSA researchers to conduct experiments on tasks of varying difficulty levels and financial practitioners to analyze market conditions at different event granularities. Additionally, our dataset can also support existing FSA tasks. By omitting event classification, our task can be simplified to a purely entity-level FSA task.

\subsection{Relatedness to ABSA Task} \label{Section 2.3}

Here, we discuss the similarities and differences between our EFSA task and the ABSA task. 

Recall that ABSA aims to identify sentiment elements related to a specific text, which could either be single or multiple elements, including the dependency relations among them~\citep{zhang2022survey,yu2021making}. These sentiment elements comprise aspect terms, sometimes aspect categories, opinion terms, and aspect-level sentiment. Among them, aspect and opinion terms are specific text spans within a sentence, and identifying them is an extraction task. Determining aspect categories and sentiments are usually classification tasks. 

Although EFSA and ABSA appear to be similar in form, as they both involve extracting the sentiment elements from the original text, i.e. companies and aspect terms, and identifying sentiment-related phrases~(events or opinion terms), and finally determining sentiment, EFSA presents two main differences.
Firstly, the events in EFSA may be extremely long and discontinuous, which results in our formulation simplifying the event extraction to a classification task. Few existing ABSA approaches can address such change without any modification. 
Secondly, the financial context makes its mapping from events to sentiments unique in EFSA. The mapping is constructed on domain expertise rather than subjective emotional expressions.
These two reasons essentially differentiate EFSA from ABSA.



The sub-tasks of EFSA exhibit a certain degree of similarity to the ABSA task. For instance, the entity-level FSA can be addressed by most current ABSA approaches. Hence, in subsequent experiments in Section~\ref{absa methods}, we benchmark several applicable ABSA baselines on our dataset.

\section{Dataset}
\begin{figure*}
    \centering
    \includegraphics[width=1\linewidth]{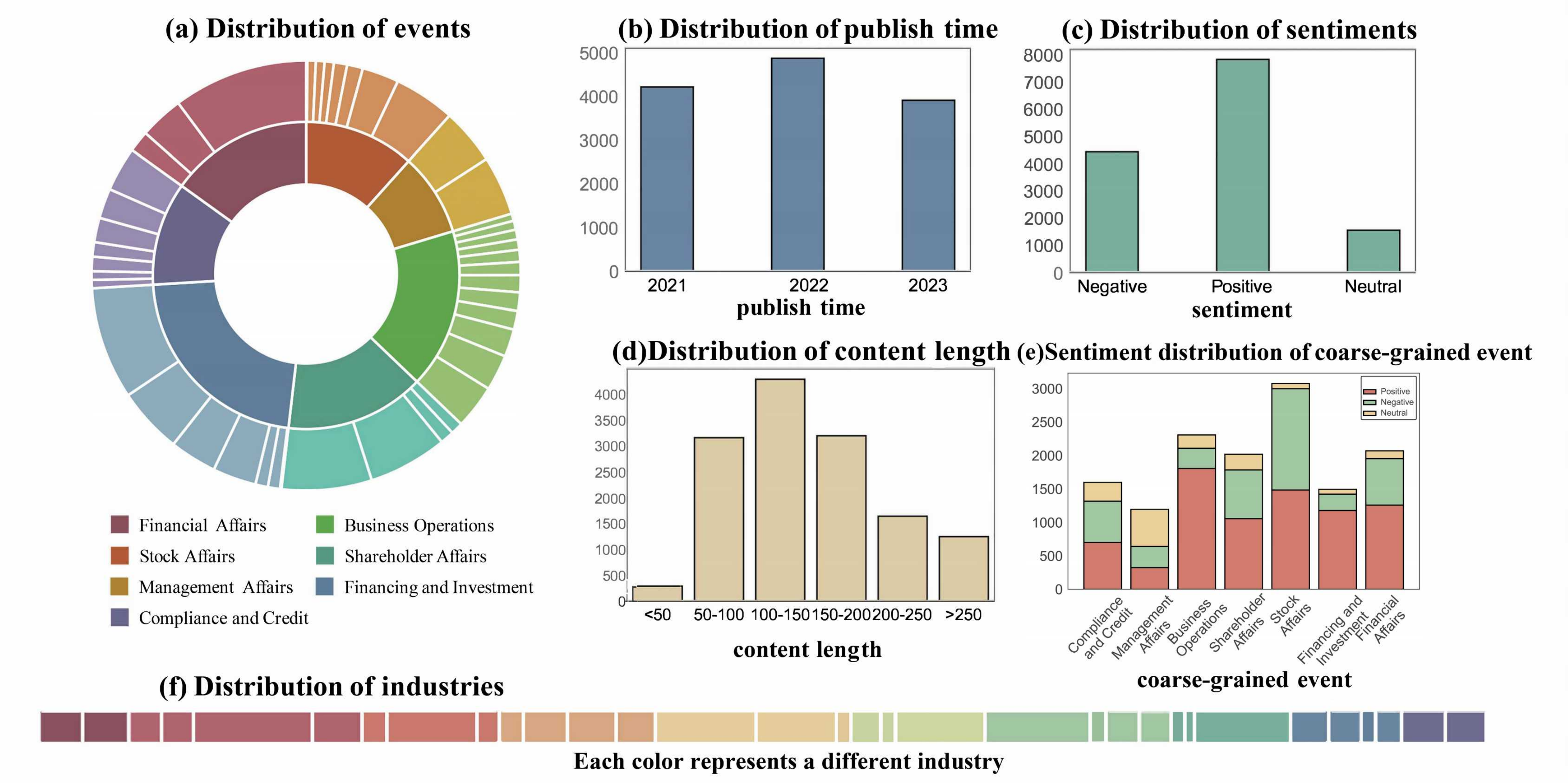}
    \caption{The statistics of data distribution of our dataset. (a) presents a nested pie chart that illustrates the distribution of event labels. The inner layer and the outer layer respectively represent the distribution of coarse- and fine-grained events; the dark and light shades of the same color represent the coarse-grained event and its subdivided fine-grained events. (b), (c), (d), and (e) present the distribution among different publish times, overall sentiments, news body length, and sentiment distribution of each coarse-grained event. (f) presents the distribution of industries, where various colors denote $32$ distinct industries. The size of the color blocks in (a) and (f) represents the data size.} 
    \label{fig:2}
\end{figure*}

Financial sentiment indicators are categorized into market-derived sentiment and human-annotated sentiments~\citep{luo2018beyond}. The market-derived sentiment is estimated based on market dynamics such as stock price changes and trading volume, potentially incorporating noise from other sources~\citep{fei2023reasoning}. 
Therefore, we employed manual labeling conducted by financial experts to ensure the creation of a high-quality dataset. Specifically, the data construction process is elaborated in detail from the following three sections: data collection, data annotation, and data distribution.

\subsection{Data Collection} 
We collected over $50,000$ financial news articles from various  reputable Chinese news outlets from their publicly accessible websites. Each collected article in the dataset includes several elements: URL, publish time, title, and news body. To control the task's complexity, we limited our selection to a news body comprising no more than $300$ Chinese characters. Based on the original data, we manually conducted data cleaning and filtering, retaining only high-quality data. Consequently, a total of $12,160$ articles were earmarked for annotation. 

\subsection{Data Annotation} 

\textbf{Labeling System.}
To address the complex spectrum of event types in financial news, we developed a detailed event taxonomy by professional financial practitioners. This taxonomy comprises seven coarse-grained categories: \textit{Financial Affairs, Shareholder Affairs, Stock Affairs, Compliance and Credit, Management Affairs, Business Operations} and \textit{Financing and Investment}, and further extends to $43$ fine-grained categories. The complete event taxonomy is presented in Appendix~\ref{sec:appendix1}, serving as a labeling reference for researchers. For sentiment labels, we adopted a three-category sentiment polarity classification consisting of positive, neutral, and negative. Each company was classified into a industry referencing the Shenwan Industry Classification Standard\footnote{The Shenwan Industry Classification is a widely used industry classification standard consisting of $32$ industry categories. It was proposed by Shenwan Hongyuan Securities for financial investment management and research.
\url{https://www.swsresearch.com/institute_sw/allIndex/downloadCenter/industryType}}.

\textbf{Annotation Platform.}
To streamline the data annotation process, we developed a specialized annotation platform tailored to our task requirements. This platform presents news bodies to annotators as input, enabling them to choose specific text spans within the news article for company labels and to directly assign event labels, sentiment labels, and industry labels within the system. The screenshot of the annotation platform is shown in Appendix~\ref{sec:appendix4}.

The annotation platform enables collaborative annotation through a two-step process: labeling and reviewing. Each news article is first labeled by an annotator and then reviewed by a reviewer. If errors are found, the data is reassigned for re-annotation. In cases of disagreement, a third annotator is tasked with resolution. 
This ensures the accuracy of each data is rigorously assessed by at least two individuals.

\textbf{Labeling Evaluation Metrics.}
To ensure consistency across various annotations, we tasked multiple annotators with independently labeling the same news article set. Following previous work~\citep{hripcsak2005agreement,barnes2018multibooked}, we use different metrics to measure the inter-annotator agreement of different annotation tasks. We employ the \textit{AvgAgr} metric~\citep{wiebe2005annotating} to evaluate span extraction annotation consistency. The \textit{AvgAgr} score of~\textit{Company} is $0.65$.
For classification annotation consistency score, we employ \textit{Fleiss' Kappa}~\citep{fleiss1971measuring} to evaluate.
The \textit{Fleiss' Kappa} scores of classification annotation are as follows: \textit{fine-grained event}~($0.62$), \textit{sentiment}~($0.67$), and \textit{industry}~($0.70$).
The results, which fall between $0.61$ and $0.8$, demonstrate a substantial agreement~\citep{landis1977measurement} among different annotators.

\subsection{Data Distribution} 
We balanced the distribution of different elements in our dataset to ensure a proportional data distribution. The specific data distribution is reported as follows:

Figure~\ref{fig:2}~(a) demonstrates the substantial uniformity of coarse- and fine-grained event distribution. For each fine-grained event, there is a sufficient amount of data to support, ensuring no instances no data shortage exists for any particular category. As shown in Figure~\ref{fig:2}~(b), We balanced the distribution of years to reflect market dynamics evenly across different periods. Most of our data's publication times span from 2021 to 2023. Figure~\ref{fig:2}~(d) demonstrates the distribution of context length. We restricted the length of the news context to 300 Chinese characters. Besides, we balanced the distribution of context length to manage the complexity of the task, with the average length of the news body in our dataset being $148.5$. Figure~\ref{fig:2}~(c) and (e) demonstrate that our dataset achieves a balanced overall sentiment distribution. Additionally, it maintains a balance between positive and negative sentiments within specific coarse-grained events, since financial texts inherently tend to exhibit a sentiment bias rather than being neutral~\citep{cortis2017semeval,de2018inf}. For fine-grained events, given that certain events have specific sentiment tendencies, such as \textit{Legal Affairs} often corresponding to negative emotions, we did not balance the sentiment distribution of each fine-grained event. We also calculated the distribution of industries ~(c.f. Figure~\ref{fig:2}~(f)) to ensure our dataset comprehensively and uniformly covers a wide range of sectors.

\section{Experimental Evaluation}

In this section, we benchmark EFSA with some widely used language models. In the following part, we will introduce the benchmark methods first, then detail our proposed framework, and finally present the evaluation metrics.

\subsection{Benchmark Methods} \label{absa methods}
\begin{figure}
    \centering
    \includegraphics[width=1\linewidth]{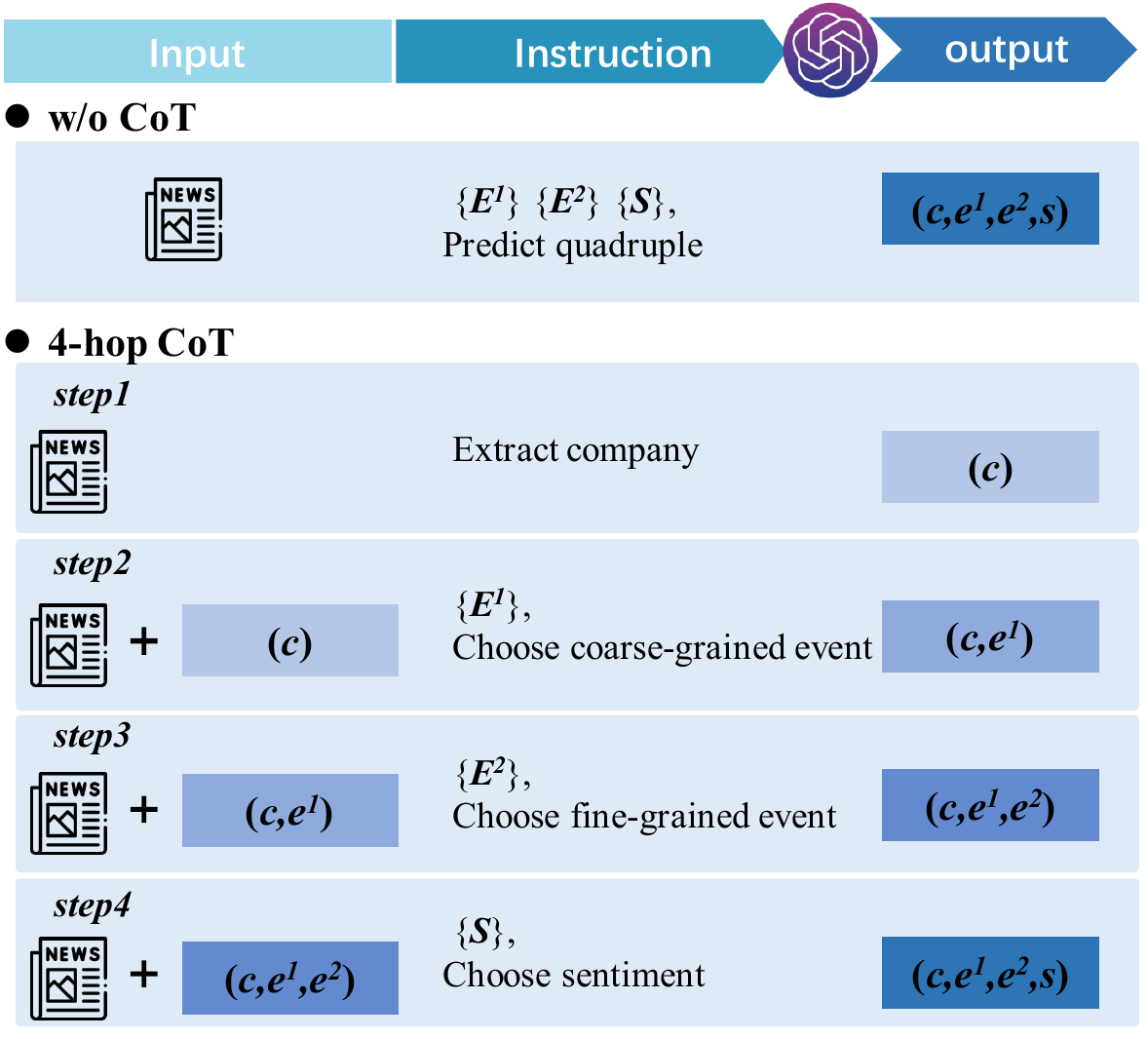}
    \caption{An illustration of the four-hop CoT framework.~\textit{E}\textsuperscript{\textit{1}},~\textit{E}\textsuperscript{\textit{2}},~\textit{S} respectively denotes the set of coarse-grained events, fine-grained events and sentiment polarities, respectively.}
    \label{fig:3}
\end{figure}
We mainly benchmark two groups of pre-trained language models, including Large Language Models~(LLMs) and Small Language Models~(SLMs).

\textbf{LLMs.} We prioritized models with strong Chinese language capabilities, selecting general domain LLMs and financial domain-specific LLMs. Particularly, we fine-tuned the open-source, deployable general domain LLMs using LoRA~\citep{hu2021lora} on our dataset. We interacted with the LLMs by constructing prompts to instruct LLMs to generate structured outputs. To ensure fairness in our benchmark tests, we use the same prompts for every LLMs under different settings, which are detailed in Appendix~\ref{sec:appendix2}.

(1) \textbf{ChatGPT}~\citep{brown2020language}. 
We interacted with ChatGPT by querying the API interface. Due to the cost constraint, we randomly selected 2,000 entries from our dataset for evaluation. 
We evaluated gpt-3.5-turbo and gpt-4-turbo-preview\footnote{The latest model interface provided by OpenAI currently points to gpt-3.5-turbo-0613 and gpt-4-0125-preview.} on this subset under both zero- and $3$-shot settings.

(2) \textbf{ChatGLM}~\citep{du2021glm} is a Chinese and English bilingual language model, constructed utilizing the General Language Model~(GLM) framework. ChatGLM-3 demonstrated superior performance on the Chinese LLM evaluation benchmark C-EVAL~\citep{huang2023c}. Moreover, we extended our analysis to the latest iteration, ChatGLM-4, by querying the API interface glm-4 provided by ZHIPU AI open platform. ChatGLM-4 stands out as one of the LLMs known for its robust Chinese language alignment capabilities. 

(3) \textbf{Llama2-Chinese} employs a Chinese instruction set for LoRA fine-tuning on Llama2-Chinese-7b~\citep{touvron2023llama} to enhance its alignment with Chinese and capabilities for Chinese dialogue.

(4) \textbf{Baichuan2}~\citep{yang2023baichuan} is a Chinese and English bilingual language model. It achieved the best performance among models of the same size on standard benchmarks (C-Eval~\citep{huang2023c}, MMLU~\citep{hendrycks2020measuring}, etc).

(5) \textbf{QwenLM}~\citep{bai2023qwen} is a Chinese and English bilingual language model. It achieved better performance than LLaMA2-70B on all tasks and outperforms GPT-3.5 on 7 out of 10 benchmarks~\citep{bai2023qwen}.

(6) \textbf{DISC-FinLLM}~\citep{chen2023disc} is a financial domain-specific LLM fine-tuned by Baichuan2-13B-Chat with LoRA on using various financial task open datasets.

(7) \textbf{Xuanyuan}~\citep{zhang2023xuanyuan} is a financial domain-specific LLM derived through incremental pre-training based on Llama2. It exhibits significant enhancements in both Chinese language proficiency and financial capabilities.

\textbf{SLMs.} 
Since \citet{zhang2021towards}'s pioneering application of generative methods to ABSA through the construction of generative paradigms, the SOTA leaderboard of ABSA has been consistently dominated by generative methods~\citep{gou2023mvp,wang2022unifiedabsa}. Given the similarities between our sub-tasks and the ABSA task, we benchmarked several advanced SLMs that achieved promising performance on ABSA. 

Leveraging the GAS~(Generative ABSA)'s framework~\citep{zhang2021towards}, we benchmarked the performance of the \textbf{mT5-large} and \textbf{BART-large-chinese}~\citep{zhang2021aspect} models on entity-level sub-task. We also benchmarked \textbf{E2E-BERT}~\citep{li2019exploiting} on this sub-task. 
We randomly split $80\%$ data for training and the left $20\%$ part is for testing. It is noted that the original training hyperparameters are used for each respective model.

\subsection{Our Framework}
Given Chain of Thought~(CoT)'s success in analyzing implicit sentiment~\citep{fei2023reasoning} and the inherent chained progression relationship between coarse- and fine-grained events in the EFSA task, we devised a four-hop CoT framework for our task. It involves four steps. 
\textbf{Step 1.} Utilize a news article as input and instruct the LLM to identify the company mentioned within the text. 
\textbf{Step 2.} Subsequently, use the news article and the identified companies as input. Instruct the LLM to choose a corresponding event from a predefined set of coarse-grained events. 
\textbf{Step 3.} Use the news article and the (company, coarse-grained-event) tuple as input. Instruct the LLM to choose a corresponding event from a predefined set of fine-grained events.
\textbf{Step 4.} Use the news article and the (company, coarse-grained event, fine-grained event) triplet as input. Instruct the LLM to choose a corresponding sentiment of the triplet.

Figure \ref{fig:3} illustrates our four-hop CoT framework. The details of the four-hop CoT prompt are displayed in Appendix~\ref{sec:appendix2}.
Based on this framework, we employed dialogue fine-tuning on deployable LLMs to evaluate effectiveness.
\begin{table*}[t]\small
    \centering
\begin{tabular}{clcccc}
\toprule
\textbf{Settings} &\textbf{Methods} & \textbf{Entity-Level FSA} & \textbf{C-EFSA} & \textbf{F-EFSA} & \textbf{EFSA} \\
\midrule
\multirow{4}{*}{\textbf{zero-shot}}&\textbf{\textit{\ \textbullet }}  \textbf{\textit{General Domain LLMs}} \\
&ChatGPT\hspace{3.2mm}  &  58.47 &  37.92 & 36.96  & 26.17  \\
&GPT-4\hspace{6.9mm}   & 60.72  &  48.48 & 45.45  &  36.10 \\ 
&GLM-4\hspace{5.5mm} &  71.25&  57.31 & 52.72 & 49.41  \\
\hdashline
\multirow{3}{*}{\textbf{3-shot}}&ChatGPT\hspace{3.2mm} &  58.92 &  39.24 &  37.13 &  27.57 \\
&GPT-4\hspace{6.9mm} & 61.38  & 51.43  & 49.53  &  39.24 \\
&GLM-4\hspace{5.5mm} & 70.39  &  56.97 & 54.90  & 50.68  \\
\hdashline
\multirow{4}{*}{\textbf{LoRa fine-tune}}&ChatGLM3-6B-Chat & 76.83 & 62.26 & 53.11 & 51.18 \\
&Baichuan2-13B-Chat & \textbf{86.41} & 71.82 & 67.79 & 67.06 \\
&Qwen-7B-Chat & 86.14 & 73.22 & 67.32 & 67.28 \\
&Llama2-Chinese-7B-Chat & 61.55 & 43.84 & 36.71 & 36.28 \\
\midrule
\multirow{3}{*}{\textbf{zero-shot}}&\textbf{\textit{\ \textbullet }}  \textbf{\textit{Financial Domain LLMs}} \\
&DISC-FinLLM  & 63.74  & 32.43 & 22.45  & 19.25  \\
&Xuanyuan &  64.15 &  22.39 &  16.41 & 7.85  \\ 
\hdashline
\multirow{2}{*}{\textbf{3-shot}}&DISC-FinLLM &  65.23 &  38.67 &  27.07 &  24.68 \\
&Xuanyuan & 63.58  & 26.70  & 17.16  &  12.39 \\
\hdashline
\multirow{2}{*}{\textbf{LoRa fine-tune}}&DISC-FinLLM & 79.19 & 56.08 & 49.16 & 46.98 \\
&Xuanyuan & - & - & - & - \\
\midrule
\multirow{4}{*}{\textbf{-}}&\textbf{\textit{\ \textbullet }}  \textbf{\textit{SLMs}} \\
&E2E-BERT & 73.77 & - & - & - \\
&GAS-T5 & 69.75 & - & - & - \\
&GAS-BART & 50.89 & - & - & - \\
\midrule
\multirow{6}{*}{\makecell[l]{\textbf{4-hop CoT}}}&\textbf{\textit{\ \textbullet }}  \textbf{\textit{Our Framework}} \\
&GPT-4   &  63.28 & 55.64  &  53.24 &  53.24 \\ 
&ChatGLM3-6B-Chat   & 78.93 & 70.61 & 65.19 & 65.19\\ 
&Baichuan2-13B-Chat   & 81.74 & 75.66 & 69.52 & 69.52 \\ 
&Qwen-7B-Chat   & 83.28 & \textbf{76.03}& \textbf{71.43} & \textbf{71.43} \\ 
&Llama2-Chinese-7b-Chat & 61.47 & 51.62 & 23.44 & 23.44 \\
\bottomrule
\end{tabular}
\caption{Comparison results on different EFSA tasks. The reported scores are F1 scores over one run. `-' denotes that the corresponding results are not available. The best results are bolded. We employ dialogue LoRa fine-tuning based on our framework, which enables the original model to achieve significant score improvements. With the 4-hop CoT framework, the F-EFSA's score is identical to the EFSA's. This is because the accurate prediction of fine-grained events is built upon the correct prediction of coarse-grained events.}
\label{tab:2}\end{table*}

\subsection{Evaluation Metrics} 
We combine the proposed framework with the aforementioned LMs on different tasks and compute the F1 scores for evaluation.
Notably, due to the presence of geographic names, stock codes, and other identifiers in the news text's company labels, a company prediction is considered correct if it is included within the gold label.  Predictions of events and sentiments must be exactly matched with the gold label to be considered correct.

\section{Experimental Results}

\subsection{Main Results and Analysis}

The results of different EFSA sub-tasks are reported in Table~\ref{tab:2}.
There are two notable observations. First, the performance across sub-tasks to the complete EFSA task exhibits a trend of declining scores, consistent with the escalating difficulty of the tasks.
Second, the overall scores of the entity-level task significantly surpass event-level tasks, which reveals the inherent complexity and challenges of our event-level tasks.
We make a more comprehensive comparison of various methods below. 

\textbf{General Domain LLMs.}
Our prior experiments demonstrate that smaller-parameter LLMs~(ChatGLM, Llama2-Chinese, Baichuan2, and QwenLM) perform poorly in zero-shot and few-shot settings~(c.f. Appendix~\ref{sec:appendix5}). This can be attributed to the inherent complexity of the EFSA task and the limited capacity of smaller models to produce structured outputs, which leads to their failure in generating the specified quadruple responce format, culminating in incorrect outputs. Fine-tuning these LLMs can significantly improve their performance scores by enhancing both domain-specific ability and capability for structured output. 

Larger-parameter LLMs~(ChatGPT, GPT-4, and GLM-4) under zero-shot settings also fail to produce satisfactory results. Few-shot settings can enhance their performance, with particularly noticeable improvements for event classification tasks and slight improvements for sentiment analysis tasks. This may be attributed to the inherent capabilities of larger-parameter LLMs in sentiment analysis tasks. Few-shot demonstrations primarily boost the LLMs' proficiency in our specialized event classification tasks.


\textbf{Financial Domain-Specific LLMs.}
Financial domain-specific LLMs are fine-tuned or pre-trained on open-source financial domain datasets based on general domain LLMs. Our benchmark results exhibit the enhanced performance of domain-specific LLMs relative to their original base models under the zero-shot settings~(DISC vs. Baichuan, Xuanyuan vs. Llama), suggesting that domain-specific customization can significantly enhance performance on previously unseen tasks within the same domain. Furthermore, domain-specific LLMs demonstrate a commendable capability for structured output, which may be attributed to other structured tasks within their training datasets. 

However, financial domain-specific LLMs do not outperform LLMs that are fine-tuned on our dataset. So we further fine-tuned financial domain-specific LLMs on our dataset. The DISC model itself is based on LoRA fine-tuning. We utilized a blend ratio of 7:3, integrating the fine-tuning weights derived from our dataset with DISC's original weights. LoRA fine-tuning is not applicable to XuanYuan, as XuanYuan is re-pretrained.  The results show that further fine-tuning can enhance the overall capabilities. Yet, it still cannot surpass the performance of their base models fine-tuned solely on our dataset. Our dataset can serve as a rich resource to facilitate the advancement of financial domain-specific LLMs.

\textbf{SLMs.}
The benchmark scores of SLMs underscore that SLMs outstrip the performance of zero-shot LLMs, affirming the conclusions presented in \citet{zhang2023sentiment}'s study: while LLMs have shown proficiency in many sentiment analysis tasks, they fall short when it comes to extracting structured sentiment and opinion information. Smaller models have learned better structured output capabilities through fine-tuning. However, due to the constraints imposed by task difficulty, the number of parameters has limited the upper potential. Consequently, the performance of SLMs does not exceed that of fine-tuned LLMs.

\textbf{Our Framework.}
Our Chain of Thought~(CoT) framework augments the efficacy of non-open-source LLMs through prompt-based adjustments alone. For open-source LLMs, the dialogue fine-tuning empowers the foundational model to realize more substantial score enhancements, particularly notable in models with initially subpar performance~(e.g., Llama). Notably, the CoT framework hurts entity-level FSA tasks. This is because the score of the (company, sentiment) tuple is directly calculated based on the tuple part of the quadruple outputted by the 4-hop CoT, the process is susceptible to error accumulation. Inaccuracies in event prediction can lead to erroneous sentiment predictions, adversely affecting the accuracy of Entity-Level FSA.


\subsection{Case Study}
\begin{figure*}
    \centering
    \includegraphics[width=1\linewidth]{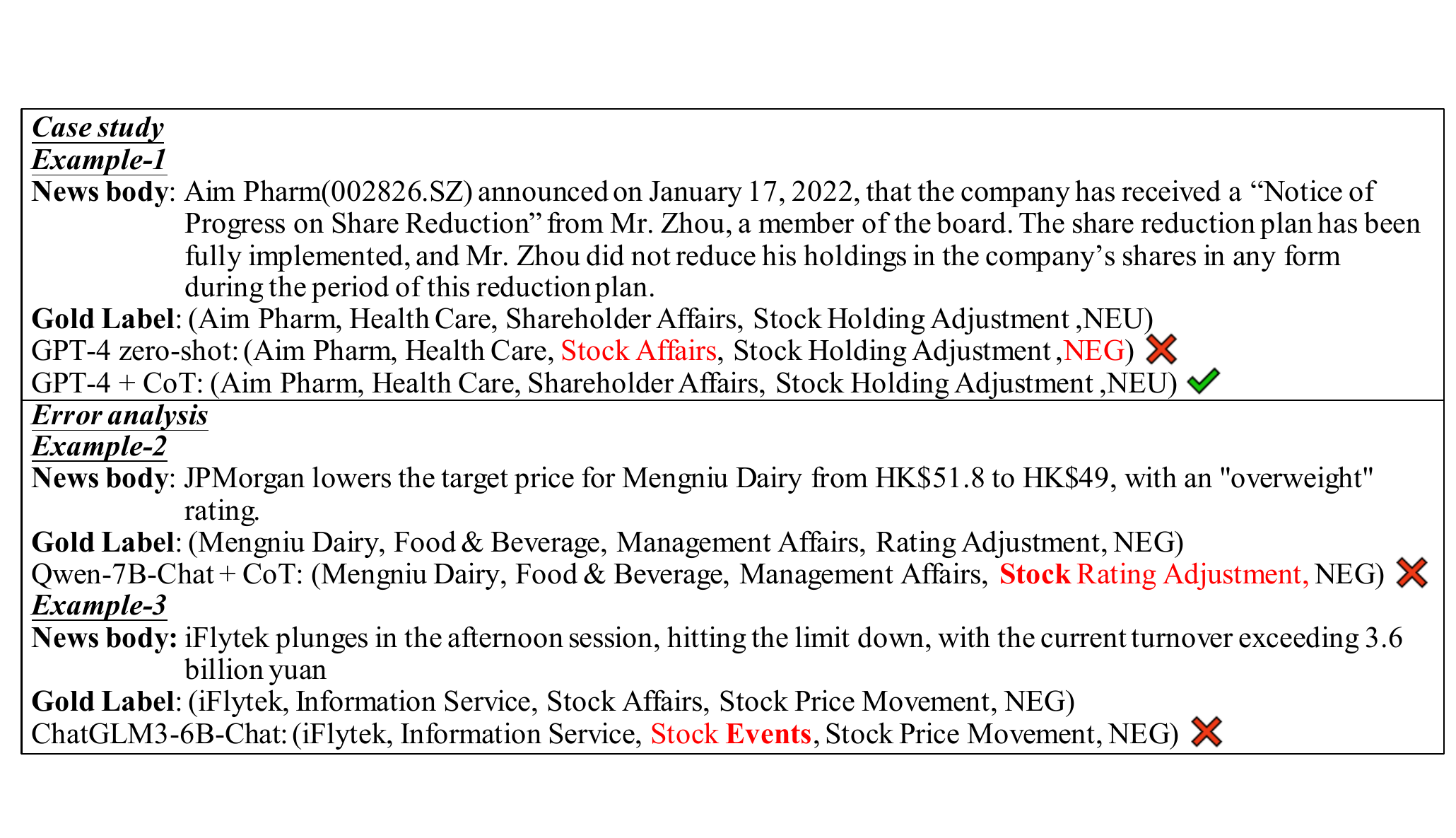}
    \caption{Examples provided include the input news text, the corresponding gold labels, and the predicted quantities. The red font denotes the incorrect part of the prediction.}
    \label{fig:4}
\end{figure*}
To better demonstrate the effectiveness of our framework, we perform a case study on GPT-4, as shown in Figure~\ref{fig:4}. 
As shown in Example 1, 
GPT-4 correctly predicted fine-grained events but made mistakes in predicting coarse-grained events, confusing the two difficult-to-distinguish categories of \textit{shareholder affairs} and \textit{stock affairs}. It is possibly due to the multiple occurrences of \textit{stock} in the text leading to misdirection. Under our CoT framework, in the second hop of reasoning from company to coarse-grained events, GPT-4 focuses more on events directly related to the company itself, thereby making the correct prediction. The CoT framework can prevent a situation where the predictions for fine-grained events are correct, yet those for coarse-grained events are not. For sentiment prediction, an inexperienced LLM can easily be misled by \textit{the company receives a notice of share reduction}. However, under the CoT framework, GPT-4 pays more attention to the sentiment of the fine-grained event~(\textit{Stock Holding Adjustment}) itself and can deduce the correct sentiment from the subsequent information that there was no actual share reduction action.

\textbf{Error Analysis.}
We observed that LLMs sometimes produce outputs beyond the defined label sets, specifically illustrated in Example-2 and -3. LLMs may alter the fixed label instead of outputting as instructed. This observation aligns with \citet{zhang2021aspect}'s previous research, highlighting the nature of the generation modeling since it does not perform ``extraction'' in the given sentence. This phenomenon is more pronounced in LLMs with smaller parameter sizes and can be mitigated through fine-tuning. Similar to instructing LLMs toward structured outputs, fine-tuning significantly enhances the ability of smaller-parameter LLMs to output as required. 

\section{Related Work}

Previous research on the FSA datasets has concentrated on sentence or document level~\citep{takala2014gold,malo2014good,cortis2017semeval,sinha2022sentfin}. This is based on an assumption that the given text conveys a single sentiment towards a certain topic. Recent FSA datasets exhibit a trend towards progressively finer granularity. However, fine-grained datasets~\citep{maia201818,tang2023finentity} mostly focus on entities and sentiments, neglecting the concern of events. 
Furthermore, the resources for fine-grained FSA datasets are still limited~\citep{du2024financial}. Following~\citet{du2024financial}, we summarize the most widely used and recent FSA benchmark datasets in Appendix~\ref{sec:appendix3}.

Event-level sentiment analysis is designed to identify user emotions on social platforms regarding current events~\citep{zhang2022enhancing}. In this paper, we broaden the scope of event-level sentiment analysis by applying it to FSA, enhancing its relevance and utility in financial contexts.

\section{Conclusion}
In this paper, we present EFSA, a novel task for Financial Sentiment Analysis~(FSA), which deepens FSA to the event level. To support this task, we constructed the largest Chinese dataset annotated for event-level FSA from a large-scale financial news corpus. We evaluated this task on widely used language models to present benchmark scores on the proposed dataset. Additionally, we designed a 4-hop reasoning prompting framework based on existing LLMs to resolve this task. Our experiments demonstrate the challenge of EFSA and the effectiveness of our proposed approach. Our work opens a new avenue in FSA and offers significant value to the entire financial domain.

\section*{Limitations} Due to budget constraints, we only conducted experiments on a subset of our dataset using LLMs that have not been open-sourced yet~(Chatgpt, GPT-4, GLM-4), on a single run. This may result in bias in our evaluation scores. Although CoT's gradual processing approach leads to more accurate and reliable results, this may increase some additional computation cost and time in solving complex FSA problems.

\section*{Ethics Statement} The collected data originated entirely from publicly accessible websites. Our datasets do not disseminate personal information and are devoid of content that could potentially harm any individual or community. The annotators are undergraduate students, which are credited as authors of the paper, and compensated in accordance with regulations. Please note that our data is intended solely as an open-source dataset resource to foster the advancement of FSA research. Any illegal use of this data is strictly prohibited.

\section*{Acknowledgements}
The research work is supported by the National Key R\&D Plan No. 2022YFC3303305. Xiang Ao is also supported by the Project of Youth Innovation Promotion Association CAS, and Beijing Nova Program 20230484430.

\bibliography{custom.bib}
\appendix

\begin{figure*}
    \centering
    \includegraphics[width=1\linewidth]{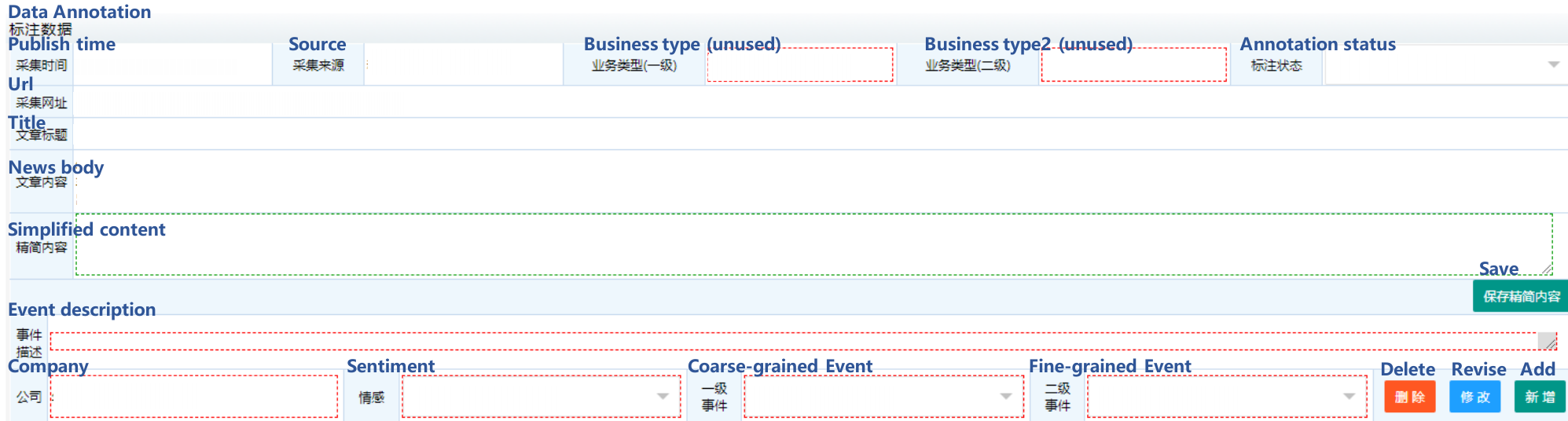}
    \caption{Screenshot of the annotation system.}
    \label{fig:5}
\end{figure*}

\begin{table*}[ht]\small
    \centering
\begin{tabular}{clcccc}
\toprule
\textbf{Settings} &\textbf{Methods} & \textbf{Entity-Level FSA} & \textbf{C-EFSA} & \textbf{F-EFSA} & \textbf{EFSA} \\
\midrule
\multirow{4}{*}{\textbf{zero-shot}}&ChatGLM3-6B-Chat & 62.07 & 19.43 & 10.14 & 8.04 \\
&Baichuan2-13B-Chat & 60.81 & 23.21 & 8.69 & 7.20 \\
&Qwen-7B-Chat & 59.36 & 20.58 & 8.15 & 5.77 \\
&Llama2-Chinese-7B-Chat & 12.62 & 1.99 & 0.43 & 0.38 \\
\hdashline
\multirow{4}{*}{\textbf{3-shot}}&ChatGLM3-6B-Chat & 66.66 & 27.87 & 16.90 & 13.86 \\
&Baichuan2-13B-Chat & 69.16 & 27.28 & 11.83 & 10.09 \\
&Qwen-7B-Chat & 63.16 & 24.55 & 16.76 & 14.60 \\
&Llama2-Chinese-7B-Chat & 41.59 & 15.15 & 6.22 & 4.96 \\
\bottomrule
\end{tabular}
\caption{Results of smaller-parameter LLMs in zero-shot and 3-shot setting}
\label{tab:4}\end{table*}

\section{Event Taxonomy}
\label{sec:appendix1}
The event taxonomy is translated from Chinese; for more accurate and detailed information, please refer to our open-source website.

\textbf{Financial Affairs}\\- Profit Announcement\\
- Profit Forecast\\
- Other Financial Affairs\\

\textbf{Shareholder Affairs}\\
- Stock Holding Adjustment\\
- Shareholder Pledge\\
- Release of Pledge\\
- Other Shareholder Affairs\\

\textbf{Stock Affairs}\\
- Stock Price Movement\\
- Equity Incentives \& Employee Stock Ownership Plans\\
- Stock Dividend\\
- Stock Buyback\\
- Stock Status\\
- Restricted Shares Release\\
- Other Stock Affairs\\

\textbf{Business Operations}\\
- Product Dynamics\\
- Capacity Changes\\
- Initiating Cooperation\\
- Technical Quality Control, Qualification Changes\\
- Government Subsidies\\
- New Company Establishment\\
- Institutional Research\\
- Intellectual Property\\
- Sales, Market Share Changes\\
- Project Bidding\\
- Project Dynamics\\
- Other Business Operations Affairs\\

\textbf{Compliance and Credit}\\
- Company Litigation\\
- Rating Adjustment\\
- Legal Affairs\\
- Clarification Announcements\\
- Regulatory Inquiries\\
- Case Investigations\\
- Administrative Penalties\\
- Other Compliance and Credit Affairs\\

\textbf{Management Affairs}\\
- Employee Dynamics\\
- Directors, Supervisors, and Senior Executives Dynamics\\

\textbf{Financing and Investment}\\
- Company Listing\\
- Mergers and Acquisitions\\
- Investment Events\\
- Stock Issuance\\
- Financing and Margin Trading\\
- Capital Flows\\
- Other Financing and Investment Affairs\\

\section{Annotation Platform}
\label{sec:appendix4}
Figure~\ref{fig:5} presents a screenshot of the annotation system, illustrating its primary functions.

\section{Prompt}
\label{sec:appendix2}
The prompts are translated from Chinese; for more accurate and detailed information please refer to our open-source website.

\textbf{Quintuple Instruction Prompt}

Assuming you are a fine-grained sentiment analysis model in the finance domain, I will give you a list of primary events, a list of secondary events, a list of sentiment polarities, and some related financial news. Please analyze which company's event is mentioned in this financial news, then determine which primary event this event belongs to, further determine the secondary event based on the primary event, and finally identify the event's sentiment polarity.

Primary Event List: [`Financial Affairs', `Shareholder Affairs', `Stock Affairs', `Management Affairs', `Compliance and Credit', `Business Operations', `Financing and Investment'].

Secondary Event List: [`Profit Announcement', `Profit Forecast', `Other Financial Affairs', `Stock Holding Adjustment', `Shareholder Pledge', `Release of Pledge', `Other Shareholder Affairs', `Stock Price Movement', `Stock Status', `Restricted Shares Release', `Stock Buyback', `Equity Incentives \& Employee Stock Ownership Plans', `Restricted Stock Release', `Stock Dividend', `Other Stock Affairs', `Directors, Supervisors, and Senior Executives Dynamics', `Employee Dynamics', `Regulatory Inquiries', `Company Litigation', `Case Investigations', `Administrative Penalties', `Clarification Announcements', `Legal Affairs', `Rating Adjustment', `Other Compliance and Credit Affairs', `Project Bidding', `Other Business Operations Affairs', `Initiating Cooperation', `New Company Establishment', `Sales, Market Share Changes', `Intellectual Property', `Technical Quality Control, Qualification Changes', `Government Subsidies', `Institutional Research', `Capacity Changes', `Project Dynamics', `Product Dynamics', `Capital Flows', `Investment Events', `Financing and Margin Trading', `Company Listing', `Mergers and Acquisitions', `Stock Issuance', `Other Financing and Investment Affairs'].

Sentiment Polarity List: [`Positive', `Negative', `Neutral'].

Please answer in the form of a list of quadruples [(Company Name, Primary Event, Secondary Event, Sentiment Polarity)].

\textbf{CoT Instruction Prompt}
\begin{table*}[t]\small
    \centering
    \begin{tabular}{lcll} 
    \toprule
         \textbf{Benchmark Dataset}&  \textbf{Entries number}&  \textbf{Annotation Type}& \textbf{Data source}\\ 
    \midrule
 \makecell[l]{Topic-Specific \\Sentiment Analysis\\ \citep{takala2014gold}}& 297& \makecell[l]{Document-level \\(Topic)} &News\\ 
 \midrule
         \makecell[l]{PhraseBank \\ \citep{malo2014good} }&  4,846&  \makecell[l]{Sentence-Level Polarity \\(sentiment polarity)}\newline& News Headlines\\ \hline 
         \makecell[l]{SemEval 2017 Task 5\\ \citep{cortis2017semeval}}&  2,836&  \makecell[l]{Sentence-Level \\(entity, sentiment score)}& \makecell[l]{News headlines \\and Posts} \\ \hline 
         \makecell[l]{SEntFiN\\ \citep{sinha2022sentfin}}&  10,753&  \makecell[l]{Sentence-Level \\(sentiment polarity)}& News headlines\\\hline 
        \makecell[l]{FiQA Task 1\\ \citep{maia201818}}& 1,173& \makecell[l]{Aspect-Level \\(entity, aspect, sentiment score)}&News\\ 
        \midrule
         \makecell[l]{FinEntity\\ \citep{tang2023finentity} }&  979&  \makecell[l]{Aspect-Level \\(entity, sentiment polarity)} & News\\ 
        \midrule
        \makecell[l]{Ours }& 12,160& \makecell[l]{Event-Level \\ (company, industry,\\coarse-grained event, \\fine-grained event,\\sentiment polarity)}&News\\ 
        \bottomrule
    \end{tabular}
    \caption{FSA Benchmark Datasets.}
    \label{tab:3}
\end{table*}

\textbf{Step~1}
Assuming you are a fine-grained sentiment analysis model in the finance domain, I will give you a piece of financial news, and you will determine the events of which companies are mentioned.

Financial news as follows: [context]

Which company's event is described in the above financial news? Only answer with the company name, if there are multiple company names, separate them with commas. Do not add extra information.

\textbf{Step~2}
Assuming you are a fine-grained sentiment analysis model in the finance domain, I will give you a piece of financial news and the company name, and you will determine the primary event in the financial news for this company.

Financial news as follows: [context]

What is the primary event occurring to [response1] in the above financial news? Please select from the following list of primary events: [`Financial Affairs', `Shareholder Affairs', `Stock Affairs', `Management Affairs', `Compliance and Credit', `Business Operations', `Financing and Investment']. You must choose from the given list of primary events, output in the form of a tuple (Company Name, Primary Event). Do not add extra information.

\textbf{Step~3}
Assuming you are a fine-grained sentiment analysis model in the finance domain, I will give you a piece of financial news, the company name, and the primary event, and you will determine the secondary event for the company.

Financial news as follows: [context]

The primary event occurring to [response1] in the above financial news is [response2], please select the corresponding secondary event from the following list of secondary events. [Appropriate Secondary Event List] You must choose from the given list of secondary events, output in the form of a tuple (Company Name, Primary Event, Secondary Event). Do not add extra information.

\textbf{Step~4}
Assuming you are a fine-grained sentiment analysis model in the finance domain, I will give you a piece of financial news, the mentioned company, the primary and secondary events, and you will determine the sentiment polarity of this financial news event.

Financial news as follows: [context]

The primary event occurring to [response1] in the above financial news is [response2], and the secondary event is [response3]. 

Please select the appropriate sentiment from [`Positive', `Negative', `Neutral'], output in the form of a quadruple (Company Name, Primary Event, Secondary Event, Sentiment Polarity).

\section{Detailed Experimental Results}
\label{sec:appendix5}
We conducted evaluations of all LLMs under zero-shot and three-shot settings. The performance of smaller-parameter LLMs counts was found to be suboptimal. Consequently, we excluded these results from the primary experimental results Table~\ref{tab:2}. These detailed results are reported in Table~\ref{tab:4}.

\section{FSA Benchmark Datasets}
\label{sec:appendix3}
Table~\ref{tab:3} summarizes the most widely used and recent FSA benchmark datasets.

\end{document}